\documentclass{article}

\usepackage[preprint,nonatbib]{neurips_2021}

\usepackage[utf8]{inputenc} 
\usepackage[T1]{fontenc}    
\usepackage{hyperref}       
\usepackage{url}            
\usepackage{booktabs}       
\usepackage{amsfonts}       
\usepackage{nicefrac}       
\usepackage{microtype}      
\usepackage{xcolor}         

\usepackage{graphicx}
\usepackage{subfigure}
\usepackage{makecell}
\usepackage[justification=centering]{caption}
\usepackage{amsmath,amssymb,amsfonts}
\usepackage{float}
\usepackage{multirow}
\usepackage{epsfig}
\usepackage{algorithm}
\usepackage{algorithmic}
\usepackage{epsfig}
\usepackage{epstopdf}
\usepackage{multirow}
\usepackage{bm}
\usepackage{times}

\title{Robust Graph Data Learning via Latent Graph Convolutional Representation}

\author{%
   Bo Jiang \\
 School of Computer Science and Technology\\
 Ahui University, China\\
  \texttt{ jiangbo@ahu.edu.cn} \\
   \And
  Ziyan Zhang \\
School of Computer Science and Technology\\
 Ahui University, China\\
 \texttt{zhangziyanahu@163.com} \\
  \AND
  Bin Luo \\
  School of Computer Science and Technology\\
 Ahui University, China\\
  \texttt{ahu\_lb@163.com} \\
}

\begin{document}

\maketitle

\begin{abstract}
 Graph Convolutional Representation (GCR) has achieved impressive performance for graph data representation.
However, existing GCR is generally defined on the input fixed graph which may restrict the representation capacity and also be vulnerable to the  structural attacks and noises. To address this issue, we propose a novel Latent Graph Convolutional Representation (LatGCR) for robust graph data representation and learning. Our LatGCR is derived based on reformulating graph convolutional representation from the aspect of graph neighborhood reconstruction. Given an input graph $\textbf{A}$, LatGCR aims to generate a flexible latent graph $\widetilde{\textbf{A}}$ for graph convolutional representation which obviously enhances the representation capacity and also performs robustly w.r.t graph structural attacks and noises. Moreover, LatGCR is implemented in a self-supervised manner and thus provides a basic block for both supervised and unsupervised graph learning tasks.
Experiments on several  datasets demonstrate the effectiveness and robustness of LatGCR.
\end{abstract}

\section{Introduction}
Graph Convolutional Networks (GCNs) have been widely studied for graph data representation and learning~\cite{defferrard2016convolutional,kipf2016semi,velickovic2017graph,DGI,IDGL,GECN}.
Graph convolutional representation (GCR) is the core
operation powering GCNs.
The aim of GCR is to generate context-aware embeddings for graph nodes by aggregating the messages from their neighbors  via some differentiable aggregation functions~\cite{graphsage,NEURIPS2020_99e314b1}.
For example,
Kipf et al.~\cite{kipf2016semi} propose a graph convolution operation by exploring the first-order approximation of graph Laplacian spectral filter. 
Hamilton et al.~\cite{graphsage} present Graph Sample and Aggregate (GraphSAGE) for inductive graph representation and learning by using graph sampling and aggregating techniques. 
Petar et al.~\cite{DGI} propose Deep Graph Infomax (DGI) to learn node's representation in an unsupervised manner.
Klicpera et al.~\cite{APPNP} propose Personalized Propagation of Neural Predictions (PPNP) which combines GCN and PageRank together for graph semi-supervised learning.
Zhu et al.~\cite{RGCN} propose Robust GCN (RGCN) by adopting robust learnable Gaussian distributions for message propagation.

However, the above GCRs are generally defined on the input fixed graph which may restrict the representation capacity and also be very vulnerable to the  structural attacks and noises~\cite{mettack,prognn,NEURIPS2020_99e314b1}. To address this issue,
one kind of popular way is
to incorporate graph learning modules into GNNs' training via optimizing a general joint loss function, i.e., `graph learning loss + GNN's loss'.
For example,
Wu et al.~\cite{GCNJaccard} propose  GCN-Jaccard to denoise input graph by deleting the edges with low similarities.
Li et al.~\cite{AdaptiveGCN} propose  AdaGCN to learn an optimal graph for GNN learning.
Jiang et al.~\cite{GLCN} propose Graph Learning Convolutional Network (GLCN) by integrating graph learning module into GCN  architecture for semi-supervised learning problem. 
Yang et al.~\cite{TOGCN}
propose Topology Optimization based Graph Convolutional Networks (TO-GCN) for semi-supervised learning by jointly refining the graph structure and learning the parameters of GCN.
Jin et al.~\cite{prognn} develop Pro-GNN to adaptively train a more optimal graph for GNN's learning via designing a joint loss function.

In this paper, we propose a novel Latent Graph Convolutional Representation (LatGCR) for \emph{robust} graph data representation and learning.
Our LatGCR is derived based on reformulating GCR from the aspect of {graph neighborhood reconstruction}.
The main difference between LatGCR and previous related works is that LatGCR gives a basic block which can be used within many GNN architectures by replacing traditional graph convolution layer with LatGCR block. We will discuss the detailed differences between LatGCR and
previous related works including recent Pro-GNN~\cite{prognn} and GeCN~\cite{GECN} in \S 4.
Specifically, given an input observed graph $\textbf{A}$, LatGCR aims to generate a flexible latent graph $\widetilde{\textbf{A}}$ for GCR in a {self-supervised} manner which enhances the representation capacity and also obviously performs robustly w.r.t graph structural attacks and noises.

Overall, we summarize the main contributions of this paper as follows:
\begin{itemize}
 \item  We propose a novel self-supervised Latent Graph Convolutional Representation (LatGCR) based on the reformulation of GCR from graph neighborhood reconstruction.
 \item LatGCR can be efficiently implemented via a simple recurrent  architecture, i.e., LatGCR block, which provides a general basic block for GNNs.
\item  Based on the proposed LatGCR block,
we propose an end-to-end  LatGC neural network (LatGCN) for robust graph data representation and learning.
\end{itemize}

  Experimental results on both semi-supervised classification and unsupervised clustering tasks demonstrate the effectiveness and robustness of the proposed LatGCR and LatGCN.

\section{Revisiting Graph Convolutional Representation}

As the main aspect of Graph Convolutional Networks (GCNs) for graph data representation and learning, Graph Convolutional Representations (GCRs) have been widely studied in recent years~\cite{defferrard2016convolutional,kipf2016semi,graphsage,velickovic2017graph,NEURIPS2020_99e314b1}.
The aim of GCRs is to generate context-aware representations for graph nodes by aggregating the representations from their neighbors via some specific aggregation functions.

One popular formulation of GCR is to employ the weighted mean aggregation function for neighbor's information aggregation~\cite{graphsage,PaGCN}.
Let $G(\mathbf{A}, \mathbf{Z})$ denotes the input graph where
$\mathbf{A}\in \mathbb{R}^{n\times n}$ denotes the adjacency matrix with $\mathbf{A}_{ii}=1$ and $\mathbf{Z}=(\mathbf{z}_1, \mathbf{z}_2\cdots \mathbf{z}_n)\in \mathbb{R}^{n\times d}$ denotes the collection of node features.
Then, the weighted mean-type GCR~\cite{graphsage} can be formulated as

\begin{flalign}\label{EQ:GCN}
\mathbf{h}_{i}' = \frac{1}{\mathbf{d}_i}\sum_{j \in \mathcal{N}_i\cup i}  \mathbf{A}_{ij} \mathbf{z}_j\mathbf{W}
\end{flalign}
where $\mathbf{d}_{i}=\sum_{j\in \mathcal{N}_i\cup i}\mathbf{A}_{ij}$ and
$\mathcal{N}_i$ represents the neighbor set of node $i$.
Matrix $\mathbf{W}\in \mathbb{R}^{d\times d'}$ denotes the graph convolutional parameter which is learned adaptively based on the specific downstream task.
The output $\mathbf{H}'=(\mathbf{h}'_1, \mathbf{h}'_2\cdots \mathbf{h}'_n)\in \mathbb{R}^{n\times d'}$ provides the graph convolutional representations of graph nodes. Comparing with input $\textbf{Z}$, $\mathbf{H}'$ involves more context information encoded in $\textbf{A}$ and also provides task-relevant representations for nodes via learned $\textbf{W}$.

\section{Latent Graph Convolution Representation}

The above GCR is defined on the fixed input graph $\textbf{A}$ which may restrict the representation capacity and also has been demonstrated to be
very vulnerable to the structural attacks and noises in $\textbf{A}$~\cite{mettack,prognn,jin2020adversarial}.
To address this issue,   we present a novel Latent Graph Convolution Representation (LatGCR) for robust graph data
representation and learning.
Our LatGCR is motivated based on the reformulation of the above GCR (Eq.(\ref{EQ:GCN}))  from the aspect of neighborhood reconstruction~\cite{NEURIPS2020_99e314b1,PaGCN}.
 Specifically, Eq.(\ref{EQ:GCN}) provides the optimal solution to
the following node reconstruction problem,
\begin{equation}\label{EQ:gcn_problem}
\begin{split}
\mathbf{h}_{i}' = \frac{1}{\mathbf{d}_i}\sum_{j \in \mathcal{N}_i\cup i}  \mathbf{A}_{ij} \mathbf{z}_j\mathbf{W}
=\arg\min_{\mathbf{h}_i} \sum_{j\in \mathcal{N}_i\cup i} \mathbf{{A}}_{ij}\| \mathbf{h}_i - \mathbf{z}_j\textbf{W}\|^2
\end{split}
\end{equation}
where $\|\cdot\|$ denotes Frobenius norm function. 

\subsection{LatGCR model formulation}

Let $G(\mathbf{A}, \mathbf{Z})$ be the input observed graph with adjacency matrix $\mathbf{A}\in \mathbb{R}^{n\times n}$ ($\textbf{A}_{ii}=1$) and node features $\mathbf{Z}\in \mathbb{R}^{n\times d}$.
The aim of LatGCR is to estimate a latent and flexible graph $\widetilde{\mathbf{A}}$ to better support GCR.
Based on the reformulation of GCR (Eq.(\ref{EQ:gcn_problem})), we present our LatGCR model by integrating
\emph{latent graph estimation} and \emph{graph convolutional representation} jointly as
\begin{align}\label{EQ:goc_problem}
\{{\widetilde{\mathbf{A}}}',\mathbf{H}'\}
& = \mathop{\arg\min}_{\widetilde{\mathbf{A}},\mathbf{H}} \,
\big\|\mathbf{A}-\widetilde{\mathbf{A}}\big\|^2 + \lambda
\sum_i \sum_{j\in \mathcal{N}_i\cup i} \widetilde{\mathbf{A}}_{ij}\big\| \mathbf{h}_i - \mathbf{z}_j\mathbf{W}\big\|^2 \\
&  s.t. \ \ \ \ \widetilde{\mathbf{A}}_{ij} \geq 0 \nonumber
\end{align}
 where $\widetilde{\mathbf{A}}'$ denotes the estimated latent graph and $\mathbf{H}'=(\mathbf{h}'_1, \mathbf{h}'_2\cdots \mathbf{h}'_n)\in \mathbb{R}^{n\times d'}$ denotes the output Latent GCRs for graph nodes.
 Matrix $\textbf{W}\in \mathbb{R}^{d\times d'}$ denotes the graph convolutional parameter and parameter $\lambda>0$ is the trade-off hyper-parameter  balancing two terms.
 The first term in Eq.(3) represents the latent graph estimation/reconstruction while the second term denotes the GCR.
Note that, when $\lambda \rightarrow 0$, the first term is penalized very largely and we can have $\widetilde{\mathbf{A}}'=\textbf{A}$. In this case,  LatGCR degenerates to standard GCR (Eq.(\ref{EQ:GCN})).
Overall, there are four main aspects of the above LatGCR model.

\begin{itemize}
\item \textbf{Self-supervised joint learning:} In LatGCR, the estimation of latent graph is conducted in a \emph{self-supervised} way. LatGCR conducts both latent graph generation and GCR jointly to boost their respective representation ability. Therefore, LatGCR can be potentially used in both supervised and unsupervised learning tasks.
  \item \textbf{Robustness:} When $\mathbf{A}$ contains some noises/errors, i.e., $\mathbf{A}=\widetilde{\mathbf{A}}+\mathbf{E}$ where $\textbf{E}$ denotes the noises/errors.
   Then Eq.(\ref{EQ:goc_problem}) can be re-formulated as
\begin{align}\label{goc_analysis}
\{{\widetilde{\mathbf{A}}}',\mathbf{H}', \mathbf{E}'\}
&= \mathop{\arg\min}_{\widetilde{\mathbf{A}},\mathbf{H},\mathbf{E}} \,
\|\mathbf{E}\|^2 +
\lambda \sum_i\sum_{j\in \mathcal{N}_i\cup i} \widetilde{\mathbf{A}}_{ij}\big\| \mathbf{h}_i - \mathbf{z}_j\mathbf{W}\big\|^2 \nonumber\\
s.t. \ \ \ \mathbf{A}&= \widetilde{\mathbf{A}} +\mathbf{E},\widetilde{\mathbf{A}}_{ij} \geq 0
\end{align}
That is, LatGCR acts as recovering/generating a latent `clear' graph $\widetilde{\mathbf{A}}$ from the input noisy graph $\mathbf{A}$ for GCR and thus performs robustly w.r.t. graph noises and attacks. This is one important property of LatGCR and will be validated in Experiments in detail.
  \item \textbf{Sparsity:}
  The estimated graph $\widetilde{\mathbf{A}}'$ inherits the same sparse pattern from input graph $\textbf{A}$, i.e., we can easily prove that
  if $\mathbf{A}_{ij}=0$, then we have $\widetilde{\mathbf{A}}'_{ij}=0$, as also seen from Eq.(5) below.
  \item \textbf{Efficient implementation:} In LatGCR, both graph estimation and graph convolutional representation are implemented via simple one-step update rules which thus can be computed very efficiently, as discussed in \S 3.2.

\end{itemize}

\subsection{LatGCR implementation.}
The optimum ${\widetilde{\mathbf{A}}}'$ and $\mathbf{H}'$ can be obtained via a simple update algorithm which {alternatively} conducts the following Latent Graph Estimation (LGE) and Graph Convolutional Representation (GCR) steps.

  \textbf{LGE-step}: Solving $\widetilde{\mathbf{A}}$ while fixing $\mathbf{H}$, the problem becomes
\begin{align}
\widetilde{\mathbf{A}}' = &\mathop{\arg\min}_{\widetilde{\mathbf{A}}}\big\|\mathbf{A}-\widetilde{\mathbf{A}}\big\|^2 +\lambda\sum_i\sum_{j\in \mathcal{N}_i\cup i}
 \widetilde{\mathbf{A}}_{ij}\big\| \mathbf{h}_i - \mathbf{z}_j\mathbf{W}\big\|^2
\ \ \ \ s.t. \ \ \ \  \widetilde{\mathbf{A}}_{ij} \geq 0 \nonumber
\end{align}
which is equivalent to
\begin{align}
{\widetilde{\mathbf{A}}}_{ij}'& =  \mathop{\arg\min}_{\widetilde{\mathbf{A}}_{ij}}\, \Big(\big(\mathbf{A}_{ij}-\frac{\lambda}{2}\|\mathbf{h}_i- \mathbf{z}_j\mathbf{W}\|^2\big)-\widetilde{\mathbf{A}}_{ij}\Big)^2 \  \  \  \  \ s.t.\ \ \ \  \widetilde{\mathbf{A}}_{ij} \geq 0 \nonumber
\end{align}
It has a simple closed-form solution which is given as 
\begin{align}\label{EQ:solveS}
\widetilde{\mathbf{A}}'_{ij} =& \max\Big\{\big(\mathbf{A}_{ij} - \frac{\lambda}{2}\|\mathbf{h}_i- \mathbf{z}_j\mathbf{W}\|^2\big), 0\Big\}
\end{align}

\textbf{GCR-step}: Solving $\mathbf{H}$ while fixing $\widetilde{\mathbf{A}}$, the problem becomes
to the standard GCR Eq.(\ref{EQ:gcn_problem}). The exact optimal solution is
\begin{equation}\label{EQ:solveH}
\mathbf{H}' = \widetilde{\mathbf{D}}^{-1}\widetilde{\mathbf{A}}\mathbf{Z}\mathbf{W}
\end{equation}
where $\widetilde{\mathbf{D}}$ is the diagonal matrix with $\widetilde{\mathbf{D}}_{ii}=\sum_{j\in \mathcal{N}_i\cup i}\widetilde{\mathbf{A}}_{ij}$.

\emph{\textbf{Remark}.}
(1) As discussed before, since the optimal $\widetilde{\mathbf{A}}'$ inherits the sparse pattern of $\mathbf{A}$, 
 in implementation of Eq.(5), we only need to compute the element $\widetilde{\mathbf{A}}'_{ij}$  with  $j\in \mathcal{N}_i\cup i$ which is efficient, as further analyzed in section \textbf{Complexity analysis}.

 (2) In real implementation, instead of using Eq.(5), we use the following update rule to avoid the possible numerical issue when $\lambda$ is large, i.e.,
 \begin{align}\label{EQ:solveS}
\widetilde{\mathbf{A}}'_{ij} =& \max\Big\{\big(\mathbf{A}_{ij} - \frac{\lambda}{2}\|\mathbf{h}_i- \mathbf{z}_j\mathbf{W}\|^2\big), \epsilon \Big\}, \,\, j \in \mathcal{N}_i\cup i
\end{align}
where $\epsilon$ is a very small positive number. That is, when $\lambda$ is large enough, we have $\widetilde{\mathbf{A}}_{ij}=\epsilon$, $j \in \mathcal{N}_i\cup i$. In this case, the above GCR (Eq.(6)) becomes to the unweighed mean aggregation which also gives a reasonable solution.

\begin{figure*}[ht]
\centering
\includegraphics[width=0.9\textwidth]{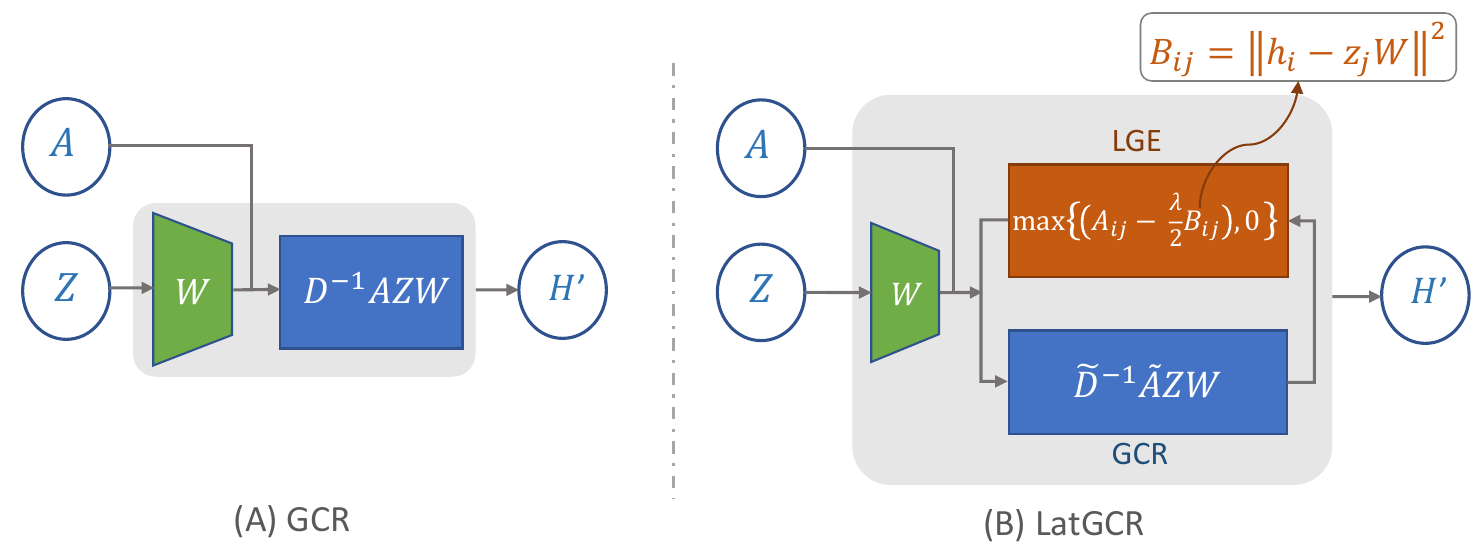}
  \caption{Architectures of  GCR and our LatGCR block.}\label{fig:module}
\end{figure*}

\textbf{LatGCR block.} From network architecture aspect, the above LatGCR implementation can be designed
via a recurrent architecture, i.e., LatGCR block, which consists of  Latent Graph Estimation (LGE) and Graph Convolutional Representation (GCR) submodules, as shown in Figure \ref{fig:module}. 
As shown in Fig. \ref{fig:module}(B), LGE and GCR are alternatively conducted in LatGCR.
On the contrary, the graph $\mathbf{A}$ is fixed in traditional GCR (shown in Fig. \ref{fig:module}(A)). Therefore, LatGCR is more flexible than GCR.
More importantly, LatGCR block performs more robustly than GCR, as demonstrated in Experiments.

\textbf{Complexity analysis.}
The whole computation complexity of the proposed LatGCR block involves three parts, i.e., linear projection, LGE and GCR.
The complexity of projection $\mathbf{Z}\mathbf{W}$ is $O(ndd')$.
The computation cost of LGE-step mainly focuses on  computing $\|\mathbf{h}_i- \mathbf{z}_j\mathbf{W}\|^2$ with  $j\in \mathcal{N}_i\cup i$ and
the complexity is $O(|\xi|d')$ where $|\xi|$ denotes edge number and $d'$ denotes dimension of feature vector $\mathbf{h}'_i$.
The complexity of GCR-step is $O(|\xi|d')$.
In summary, the whole LatGCR complexity is
$O(ndd')+O(2r|\xi|d')$ where $r$ is the recurrent time for alternatively conducting LGE and GCR and set to 3 in experiments.
Note that, LatGCR does not bring very high complexity when comparing with GCR (Eq.(1)) whose main complexity is $O(ndd')+O(|\xi|d')$.

\section{LatGC Neural Networks}

\begin{figure*}[ht]
\centering
\includegraphics[width=0.8\textwidth]{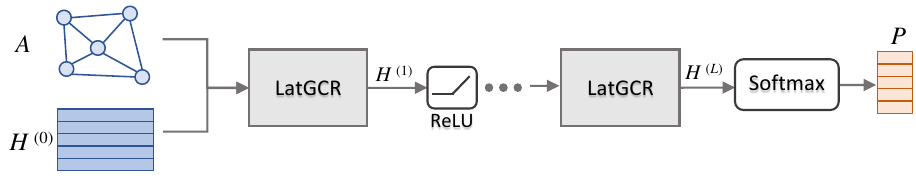}
  \caption{The architecture of LatGCN.}\label{fig:architecture}
\end{figure*}

LatGCR gives a basic block which can be used within many GNN architectures~\cite{DGI,graphsage,kipf2016semi} by replacing the traditional graph convolution layer with LatGCR module.
Here, we adopt the GNN architecture utilized in traditional GCN~\cite{kipf2016semi} and design an end-to-end multi-layer neural network architecture, named Latent Graph Convolutional Network (LatGCN) for graph data learning.
Concretely, the proposed LatGCN contains one input layer, several hidden propagation layers and one final perceptron layer, as shown in Figure \ref{fig:architecture}.
 For each hidden propagation layer, it takes features $\textbf{H}^{(l)}$ and initial graph $\textbf{A}$ as input and outputs features $\textbf{H}^{(l+1)}$ by using LatGCR module with  parameter $\textbf{W}^{(l)}$.
LatGCN can be used in many graph learning tasks.
For example, when applying LatGCN for semi-supervised node classification tasks, the last perceptron layer outputs the final label predictions $\mathbf{P}$ for all nodes.
 The convolutional parameter $\textbf{W}^{(l)}$ of each hidden layer can be learned via minimizing  the cross-entropy function on the labelled nodes, as discussed in work~\cite{kipf2016semi}.

\textbf{Comparison with related works.}
Exploiting graph learning for GNNs has been studied in recent years.
The main differences between LatGCN and previous graph learning guided GCNs including Pro-GNN~\cite{prognn}, TO-GCN~\cite{TOGCN} and GLCN~\cite{GLCN} are follows.
First, LatGCN is designed based on our proposed new LatGCR block (Figure 1 (B)).
In contrast, previous works generally incorporate graph learning into GNN's training via designing a joint loss function.
Second, LatGCR is derived based on the joint reconstruction framework (Eq.(3)) which is implemented in a self-supervised manner. This makes LatGCR be a general block which can be used within many GNN's architectures to derive various kinds of LatGCNs.

LatGCR is also different from recent GeC and GeCN~\cite{GECN}.
(1) LatGCR aims to generate  a flexible latent graph for graph convolution while GeC incorporates  neighborhood selection into graph convolution.
(2) LatGCR is derived based on neighborhood reconstruction while GeC is designed based on graph Laplacian regularization.
(3) LatGCR implements both graph estimation and GCR via simple one-step update rules while  GeC~\cite{GECN} adopts $T$-step update rules for both neighborhood selection and graph convolution operation. Thus, the implementation of LatGCR is generally more efficient than GeC~\cite{GECN}.

\section{Experiment}

To verify the effectiveness and robustness of LatGCR block and LatGCN, we test it on both semi-supervised node classification and unsupervised clustering tasks on three standard benchmark datasets, i.e., Cora, Citeseer and Pubmed~\cite{sen2008collective,prognn}.

\subsection{Semi-supervised node classification}

\subsubsection{Experimental setting}

Similar to the architecture of GCN~\cite{kipf2016semi}, LatGCN consists of one input layer, two LatGCR layers and one final  perceptron layer. The skip-connection strategy is also utilized in LatGCN, as suggested in work~\cite{kipf2016semi,7780459}.
We optimize the network weight matrices of all LatGCR modules by minimizing the cross-entropy loss function~\cite{kipf2016semi}.
For fair comparison,
we use the same attacked data setting used in work~\cite{prognn} and employ two types of attacks, i.e., Metattack~\cite{mettack} and Random Attack~\cite{jin2020adversarial}.
For Metattack~\cite{mettack},  we utilize the most destructive attack variant `Meta-Self' and
set the perturbation level from $0$ to $25\%$ with step $5\%$.
For Pubmed dataset, we use the approximate version `A-Meta-Self'  as used in work~\cite{prognn}
For Random Attack~\cite{jin2020adversarial},  we apply the variant `Add' and
set the perturbation level from $0$ to $100\%$ with step $20\%$.
Following the experimental setup in previous works~\cite{kipf2016semi,velickovic2017graph},
we set the number of units in each hidden layer to $16$
and train our LatGCN by using Adam algorithm~\cite{Adam} with learning rate of $0.001$.
The recurrent time of each LatGCR block is set to $3$ and the hyper-parameter $\lambda$ is determined based on validation set.
We provide additional experiments across different settings of parameter $\lambda$ in \S 5.3.
\begin{figure*}[ht]
\centering
\includegraphics[width=1.0\textwidth]{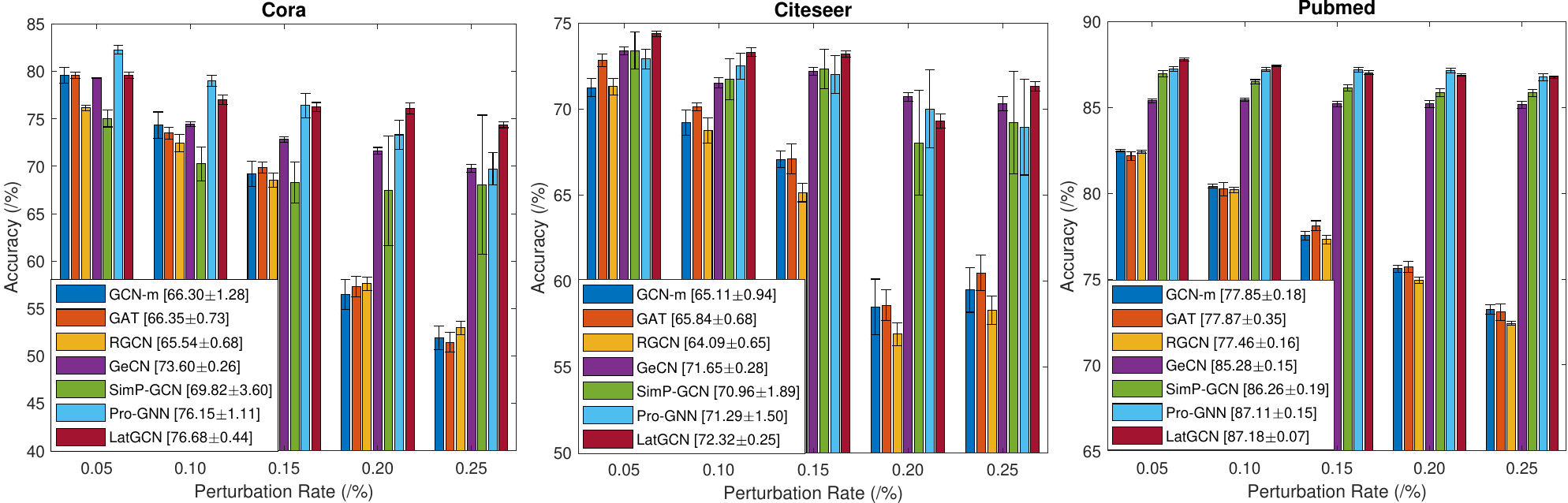}
  \caption{Semi-supervised classification performance under Metattack~\cite{mettack}.}
\label{fig:result-semi-noise-meta}
\end{figure*}
\begin{figure*}[ht]
\centering
\includegraphics[width=1.0\textwidth]{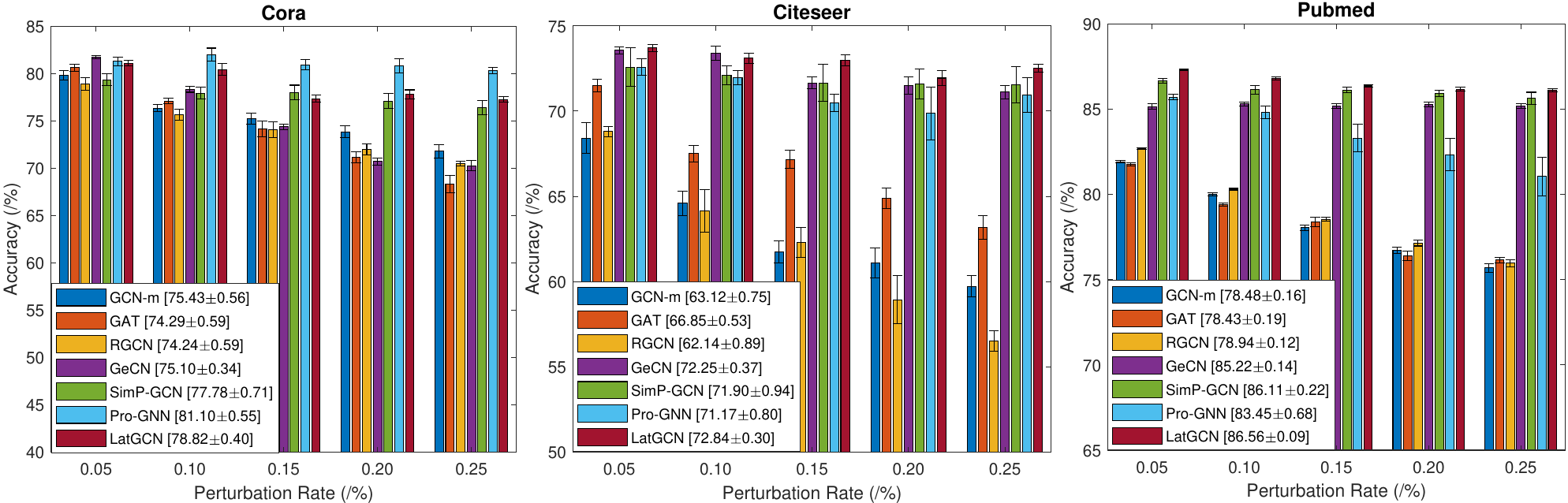}
  \caption{Semi-supervised  classification performance under Random Attack~\cite{jin2020adversarial}.}
\label{fig:result-semi-noise-random}
\end{figure*}
\subsubsection{Comparison results}

We first compare LatGCN with some traditional baseline methods including weighted mean-type GCN (GCN-m)~\cite{kipf2016semi,graphsage}, Graph Attention Networks (GAT)~\cite{velickovic2017graph}.
To demonstrate the robustness of LatGCN,
we also compare LatGCN with some recent robust GNNs including Robust Graph Convolutional Networks (RGCN)~\cite{RGCN}, SimP-GCN~\cite{SimPGCN}, Property Graph Neural Networks (Pro-GNN)~\cite{prognn} and Graph elastic Convolutional Networks (GeCN)~\cite{GECN}.

\textbf{Effectiveness analysis.}
Figure \ref{fig:result-semi-noise-meta} and \ref{fig:result-semi-noise-random} summarize the comparison results across different levels of Metattack~\cite{mettack} and Random Attack~\cite{jin2020adversarial}, respectively. 
For each attack level, the results are the average performance of 10 runs with different network initializations. 
The overall average performance of comparison methods for all attack levels are reported in the legend of each Figure. 
Here, we can observe that
(1) Traditional GCN-m~\cite{kipf2016semi,graphsage} and GAT~\cite{velickovic2017graph} are vulnerable to the structural attacks and noises. Comparing with them, LatGCN obtains obviously better performance under various attacks and noises. This clearly demonstrates the effectiveness and robustness of the proposed LatGCR module for robust graph data learning.
(2) LatGCN obtains better performance than some recent robust GNN methods including RGCN~\cite{RGCN} and SimP-GCN~\cite{SimPGCN}, which indicates
the more robustness of the proposed LatGCN w.r.t graph structural attacks and noises.
(3) Comparing with some recent graph learning guided GNNs including Pro-GNN~\cite{prognn} and GeCN~\cite{GECN}, LatGCN generally obtains the best average learning performance on attacked graph data. This further demonstrates the effectiveness of the proposed self-supervised latent graph estimation in LatGCR for noisy graph data representation. 
\begin{figure*}[ht]
\centering
\includegraphics[width=1.0\textwidth]{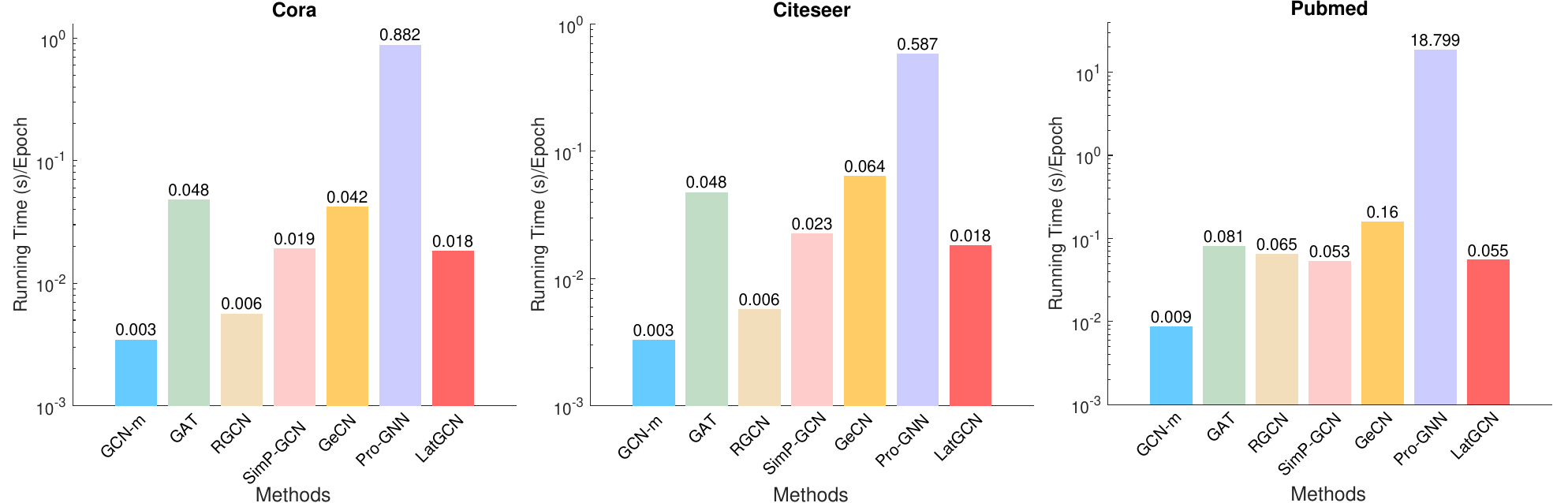}
  \caption{The empirical average running time in each epoch of different methods.}\label{fig:time}
\end{figure*}

\textbf{Efficiency analysis.}
Figure \ref{fig:time} shows the average running time of each epoch in training LatGCN on semi-supervised node classification tasks on the attacked datasets used in work~\cite{prognn}. All methods are implemented by PyTorch on NVIDIA A6000.
We can note that
(1) The methods using fixed graph GNNs, such as GCN-m, SimP-GCN~\cite{SimPGCN} and RGCN~\cite{RGCN}, generally run faster than graph learning guided GNN methods.
(2) Our LatGCN performs obviously faster than most of graph learning guided methods including GAT~\cite{velickovic2017graph}, GeCN~\cite{GECN} and Pro-GNN~\cite{prognn}, especially on the larger dataset Pubmed~\cite{sen2008collective}.
It demonstrates the efficiency of the proposed LatGCN on conducting robust graph data learning.

\subsection{Unsupervised clustering}

To evaluate the effectiveness of the proposed self-supervised LatGCR, we further test it on
unsupervised clustering tasks.
Following the experimental setting in previous work~\cite{AGCN},
we first use Singular Value Decomposition (SVD) to replace projection step to obtain low-dimensional embeddings  for graph nodes.
Then, we utilize LatGCR to obtain context-aware representations for graph nodes and employ K-means clustering algorithm~\cite{cluster} to obtain the final clustering results~\cite{AGCN}.
We set the recurrent time and parameter $\lambda$ to $1$ and $0.03$ respectively.
Similar to work~\cite{AGCN}, we adopt three widely used performance measurements~\cite{cluster}, i.e., clustering accuracy (Acc), normalized mutual information (NMI) and macro F1-score (F1) for evaluation.

We compare our LatGCR with some other popular clustering approaches including Graph Variational Autoencoder (VGAE)~\cite{GAE}, Marginalized Graph Autoencoder (MGAE)~\cite{MGAE}, Adversarially Regularized Variational Graph Autoencoder (ARVGE)~\cite{ARGE} and Attributed Graph Clustering (AGC) ~\cite{AGCN}.
Table \ref{tab:result-cluster} summarizes the comparison results on all original datasets~\cite{sen2008collective}.
The results of these comparison methods have been reported in work~\cite{AGCN} and we use them directly.
From Table 1, we can note that comparing with some other clustering methods, the proposed LatGCR can obtain the best average performance in most cases,
which further indicates the effectiveness of the proposed LatGCR on conducting unsupervised clustering tasks.
\begin{table}[!htpb]
\centering
\caption{\upshape Comparison results on clustering task. The best results are marked by bold.}
\label{tab:result-cluster}
\small
\renewcommand\arraystretch{1.1}
\begin{tabular}{c|ccc|ccc|ccc}
  \hline
  \hline
   \multirow{2}{*}{Method}& \multicolumn{3}{c|}{Cora}& \multicolumn{3}{c|}{Citeseer}& \multicolumn{3}{c}{Pubmed}\\
  \cline{2-10}
 &Acc\%&NMI\%&F1\% &Acc\%&NMI\%&F1\% &Acc\%&NMI\%&F1\%   \\
  \hline
   VGAE      & 55.95&38.45&41.50&44.38&22.71&31.88&65.48&25.09&50.95\\
   MGAE      & 63.43&45.57&38.01&63.56&39.75&39.49&43.88&8.16&41.98\\
   ARVGE     & 63.80&45.00&62.70&54.40&26.10&52.90&58.22&20.62&23.04\\
   AGC       & 68.92&\textbf{53.68}&\textbf{65.61}&67.00&41.13&62.48&69.78&31.59&68.72\\
   \hline
   LatGCR    & \textbf{69.19}&53.48&65.50 & \textbf{67.83}&\textbf{42.07}&\textbf{63.31} & \textbf{70.52}&\textbf{32.77}&\textbf{69.71}\\
  \hline
  \hline
\end{tabular}
\end{table}
\subsection{Model analysis}

\subsubsection{Visualization results}

\begin{figure}[htpb]
\centering
\subfigure[GCN-m result on Cora dataset]{\includegraphics[width=2.5in]{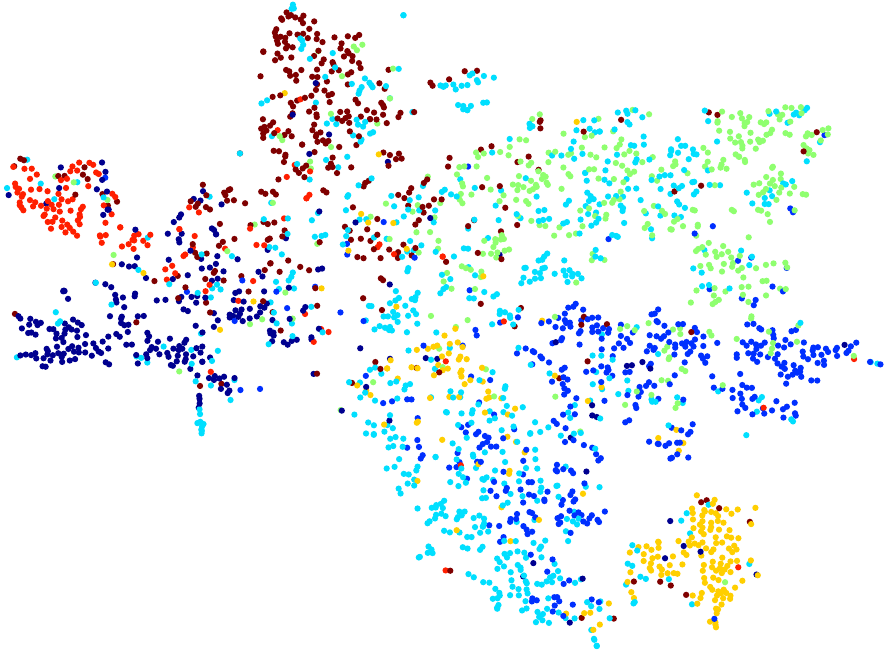}}
\subfigure[LatGCN result on Cora dataset]{\includegraphics[width=2.5in]{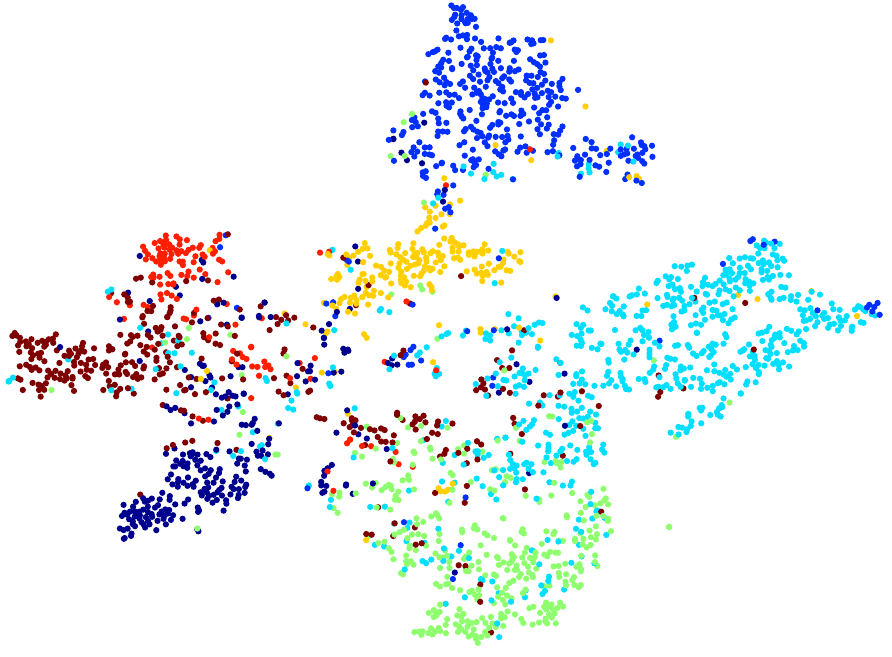}}
\subfigure[GCN-m result on Citeseer dataset]{\includegraphics[width=2.5in]{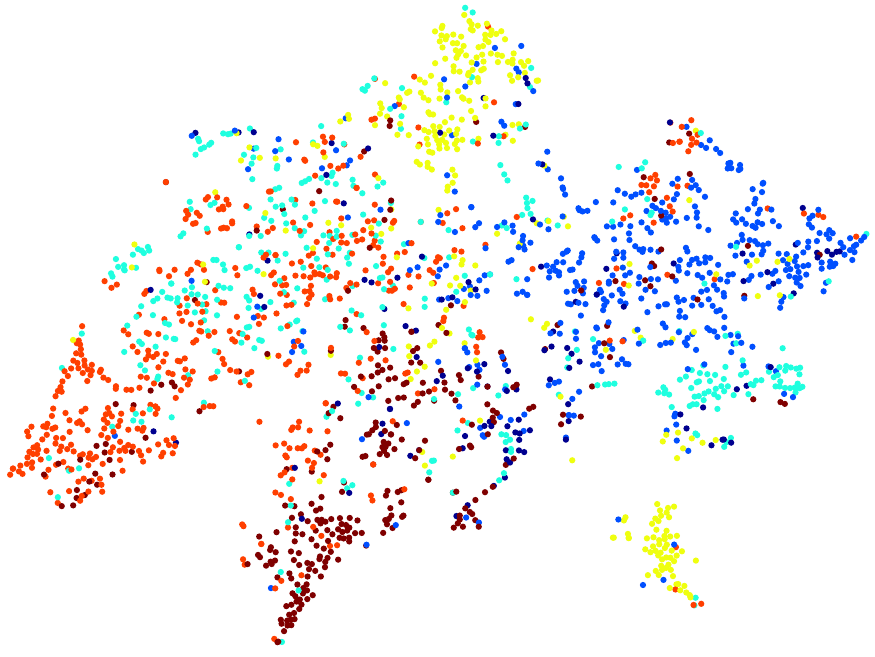}}
\subfigure[LatGCN result on Citeseer dataset]{\includegraphics[width=2.5in]{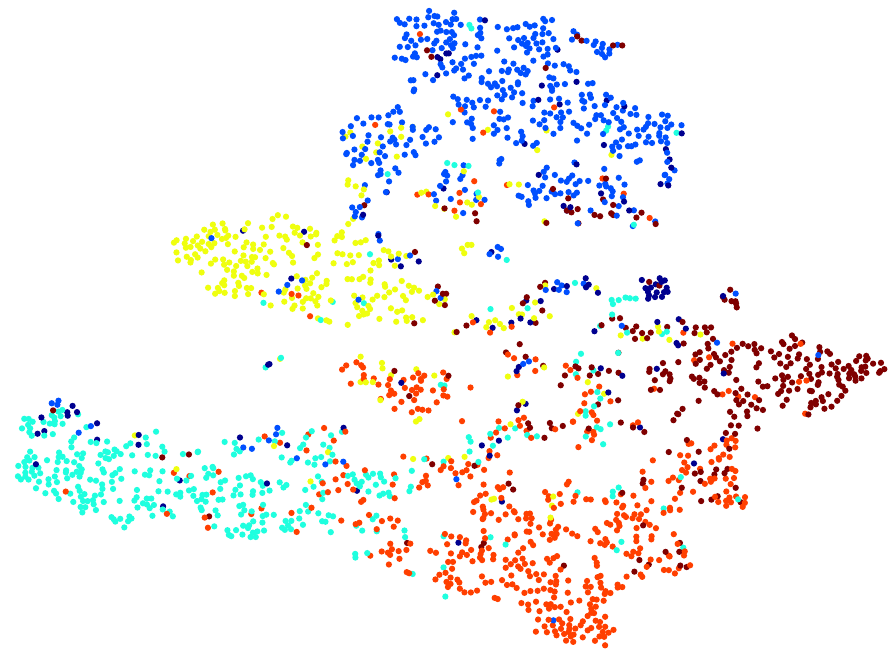}}
\caption{2D t-SNE~\cite{tsne} visualizations of the feature map output from the first layer of GCN-m and LatGCN. Different colors denote different classes.}
\label{fig:demo}
\end{figure}

To demonstrate the effectiveness of proposed LatGCR, we utilize 2D t-SNE~\cite{tsne} visualization to show the feature representation ability of LatGCR comparing with baseline method GCN-m~\cite{graphsage,kipf2016semi}.
Figure~\ref{fig:demo} shows 2D t-SNE~\cite{tsne} visualization results of the feature maps output by the first hidden layer of GCN-m and LatGCN on Cora and  Citeseer~\cite{sen2008collective,prognn} datasets under $0.25$\% Metattack~\cite{mettack}. Different colors denote different classes.
One can note that, LatGCR obtains clearer and compacter embeddings than baseline method GCN-m, which intuitively demonstrates that the proposed LatGCR can obtain more robust and discriminative feature representations for graph data with structural attacks.

\begin{table}[!htpb]
\centering
\caption{\upshape Results of LatGCN with different parameter $\lambda$ values.}
\label{tab:lambda}
\renewcommand\arraystretch{1.3}
\begin{tabular}{c|ccccccc}
  \hline
  \hline
  $\lambda$  & 0.05 & 0.1& 0.5 & 1   &5    &   10&20\\
  \hline
   Cora      &76.06&77.11&76.36&75.96&72.79&70.82&67.25 \\
   Citeseer  &71.09&71.27&71.68&71.5&72.16&70.44&69.73\\
   Pubmed    &84.87&85.36&86.03&86.07&86.42&86.66&86.40 \\
  \hline
  \hline
\end{tabular}
\end{table}
\subsubsection{Parameters analysis}

One main hyper-parameter in LatGCR is the balanced parameter $\lambda$ (Eq.(3)).
Table \ref{tab:lambda} shows the semi-supervised classification results of LatGCN with different $\lambda$ values on three datasets under $100\%$ Random Attack~\cite{jin2020adversarial}.
We can see that our LatGCN can achieve relatively stable results in a certain range of parameter $\lambda$ value which indicates that LatGCR is generally insensitive to the value of hyper-parameter $\lambda$ in a certain range.

\section{Conclusion}

This paper proposes a novel Latent Graph Convolutional Representation (LatGCR) for robust graph data representation and learning.
LatGCR is proposed based on a joint reconstruction framework, i.e., graph structure reconstruction + node's feature reconstruction.
 It  can estimate a latent and flexible graph for GCR in a self-supervised way and provides a general basic block for GNNs.
The main advantage of LatGCR is that it can perform robustly w.r.t graph structural attacks and noises.
Experiments on several benchmark datasets demonstrate the effectiveness and robustness of LatGCR.
In our future work, we will extend LatGCR to address the data with multiple graphs and further apply it on some more applications and tasks, such as computer vision, recommendation, etc.

\end{document}